\documentclass[12pt]{article}
\usepackage{ttr}
\usepackage{graphicx}
\usepackage{amsmath, amssymb}
\usepackage[hidelinks]{hyperref}
\usepackage{caption}
\usepackage{enumitem}
\usepackage[numbers]{natbib}
\usepackage[most]{tcolorbox}
\usepackage{booktabs}
\usepackage{xcolor}
\usepackage{listings}
\usepackage{tikz}
\usetikzlibrary{shapes, arrows.meta, positioning}

\usepackage{textcomp}
\title{Retrieval-Augmented Reasoning \\ with Lean Language Models}
\author{Ryan Sze-Yin Chan$^{1}$, Federico Nanni$^{1}$, Tomas Lazauskas$^{1}$,\\ Rosie Wood$^{1}$, Penelope Yong$^{1}$, Lionel Tarassenko$^{2}$, \\ Mark Girolami$^{1,3}$, James Geddes$^{1}$, Andrew Duncan$^{4}$\smallskip \\
 $^1$ The Alan Turing Institute, $^2$ University of Oxford,\\$^{3}$ University of Cambridge, $^{4}$ Imperial College London\smallskip \\
 Corresponding authors:\\\smallskip  \tt\{rchan,fnanni,jgeddes\}@turing.ac.uk \\\smallskip
 \tt a.duncan@imperial.ac.uk}
\date{}
\begin{document}
\maketitle
\renewcommand{\abstractname}{\sffamily Abstract}
\begin{abstract}
    This technical report details a novel approach to combining reasoning and retrieval augmented generation (RAG) within a single, lean language model architecture. While existing RAG systems typically rely on large-scale models and external APIs, our work addresses the increasing demand for performant and privacy-preserving solutions deployable in resource-constrained or secure environments. Building on recent developments in test-time scaling and small-scale reasoning models, we develop a retrieval augmented conversational agent capable of interpreting complex, domain-specific queries using a lightweight backbone model. Our system integrates a dense retriever with fine-tuned \texttt{Qwen2.5-Instruct} models, using synthetic query generation and reasoning traces derived from frontier models (e.g., \texttt{DeepSeek-R1}) over a curated corpus—in this case, the NHS A-to-Z condition pages. We explore the impact of summarisation-based document compression, synthetic data design, and reasoning-aware fine-tuning on model performance. Evaluation against both non-reasoning and general-purpose lean models demonstrates that our domain-specific fine-tuning approach yields substantial gains in answer accuracy and consistency, approaching frontier-level performance while remaining feasible for local deployment. All implementation details and code are publicly released to support reproducibility and adaptation across domains.
\end{abstract}

\section{Introduction}

Recent efforts to improve the test-time performance of language models have shown significant promise \cite{jaech2024openai,snell2025scaling,zuo2025ttrl,liu2025can}. These approaches, particularly those that target enhancements of so-called ``reasoning'' capabilities via chain-of-thought prompting \cite{wei2022chain}, have enabled relatively small-scale models (e.g., \texttt{DeepSeek-R1} distilled models \cite{guo2025deepseek} or \texttt{s1} \cite{muennighoff2025s1}) to achieve results that are comparable to those of frontier models (e.g., OpenAI's offerings \cite{elopenai}), in specific tasks.

In parallel, work that has focused on improving the factuality and verifiability of the output of LLMs through \emph{retrieval augmented generation} (RAG) strategies has presented clear opportunities to reduce hallucinations \cite{lewis2020retrieval,gao2023retrieval,fan2024survey}, in particular when dealing with the complexities of specific domains of knowledge \cite{ram2023context,asai2024reliable}. 

The successful integration of reasoning and RAG is now widely available in tools like ChatGPT and Gemini. Given a user query, these systems can, for instance, first reason about the query and then decide to take an action---such as performing a web search or querying a tool like Google Maps---before returning a final answer. This form of reasoning and tool use is characteristic of emerging agentic AI systems \cite{anthropic2024effective_agents,wang2024survey, xi2025rise}. Alternatively, the system may begin by retrieving documents relevant to the user query and then reason over the collected evidence before responding. This second approach---retrieval followed by reasoning---will be the focus of this technical report.

While the combination of retrieval and reasoning has significantly enhanced the performance of frontier language models in general-purpose applications, such approaches encounter clear limitations in scenarios where users are unwilling or unable to share data with external entities---particularly in domains involving sensitive or private information. Even if the training data of a model is publicly available, the prompts posed by users can often contain highly proprietary or sensitive information which cannot cross organisational or national boundaries.

In these cases, it becomes necessary to deploy language models on local infrastructure, potentially within secure or air-gapped environments. To address such requirements, recent years have seen steady progress in the development of openly available large language models (e.g., \cite{bai2023qwen,touvron2023llama,team2024gemma}) alongside open-source frameworks for retrieval augmented generation\footnote{See for instance \url{https://www.llamaindex.ai/} and \url{https://www.langchain.com/}.}. More recently, small-scale reasoning models have also begun to emerge \cite{guo2025deepseek,muennighoff2025s1}. Nonetheless, the effective integration of reasoning capabilities for interpreting retrieved evidence---particularly within the constraints of lightweight or locally deployable models---remains an open research challenge. While some recent work such as ReAct \cite{yao2023react}, REPLUG \cite{shi2024replug}, and MemGPT \cite{packer2023memgpt} explore hybrid architectures for strongly integrating LLM reasoning with document retrieval, they are mostly in large, non-local model settings.


To address these limitations, this technical report presents an approach for effectively combining reasoning and retrieval augmented generation within a single, lean language model. Furthermore, we integrate this fine-tuned model into an interactive conversational system to demonstrate its applicability in downstream tasks. The resulting system is particularly well-suited for applications involving complex queries over private, domain-specific knowledge bases. In such settings, the reasoning component facilitates the interpretation and decomposition of intricate queries, while the retrieval mechanism constrains the model to verifiable information, thereby mitigating the risk of hallucinated responses. The focus on private and sensitive domains motivated our emphasis on lean language models that can be feasibly fine-tuned and deployed by small organizations or government departments, particularly in compute-constrained or secure environments.


The report is structured as follows. We begin with an overview of test-time scaling strategies and related work relevant to the task. This is followed by a detailed description of our system architecture, including implementation choices and practical guidance for reproducibility, supported by references to the accompanying codebase. We then demonstrate the application of our approach to a representative domain-specific knowledge base---the NHS A-to-Z condition webpages\footnote{Publicly available at \url{https://www.nhs.uk/health-a-to-z/conditions/}}---using a set of queries that require both retrieval and reasoning capabilities. The report concludes with a discussion of potential future enhancements. An open-source implementation of our method is available via GitHub,\footnote{\url{https://github.com/alan-turing-institute/t0-1}} enabling practitioners to apply the system to a broad range of problems involving domain-specific question answering that combines retrieval with structured reasoning.


\section{Related work}

In the following section we provide an overview of research areas relevant to this technical report. 

\subsection{Test-time scaling}

The core concept of test-time scaling involves enhancing the performance of large language models (LLMs) by increasing computational resources during inference rather than during pre-training. Prior work has demonstrated that this strategy can lead to improved performance more efficiently than increasing compute during the pre-training phase itself \cite{snell2025scaling,liu2025can}. In practice, test-time scaling refers to deploying inference-time strategies that leverage additional sampling, computation, or prompt engineering to boost the capabilities of a fixed model—without modifying its parameters through fine-tuning or reinforcement learning.

One widely used class of test-time scaling methods is \emph{parallel generation}, where a model generates multiple candidate responses and then aggregates them through selection mechanisms such as \emph{majority voting} \cite{zuo2025ttrl}, \emph{self-consistency} \cite{chen2023universal}, or \emph{best-of-$N$ sampling} \cite{brown2024large}. These techniques improve robustness and factual accuracy by exploiting diversity in the model's outputs, with selection based on heuristics or learned reward functions. Other common strategies such as beam search \cite{graves2012sequence} and Monte Carlo tree search \cite{yao2023tree}, which maintains multiple high-probability continuations of a sequence in parallel to explore more optimal generations. While such approaches typically improve likelihood, they may reduce diversity, in contrast to sampling-based methods.


A complementary family of approaches is known as sequential scaling, which involves increasing the number of intermediate reasoning steps the model takes before arriving at a final answer. The most prominent example is \emph{chain-of-thought prompting} \cite{wei2022chain}, in which models are guided to produce intermediate reasoning steps that improve performance on complex tasks. This trend has contributed to a broader anthropomorphisation of model behaviour, often described in terms of “reasoning” \cite{kambhampati2025stop}. Extensions such as \emph{tree-of-thought prompting} generalise this idea by exploring multiple reasoning paths in a branching structure, potentially with scoring and pruning mechanisms applied to select the most promising trajectory. More advanced test-time scaling methods differ in whether they assume access to a \emph{verifier}---a model or module that can score, rerank, or validate outputs. In verifier-free settings, selection relies on internal model heuristics (e.g., majority vote, self-consistency), while verifier-assisted setups may use external reward models, classifiers, or even humans to evaluate and select responses, leading to higher precision but increased complexity.

Recent models such as \texttt{DeepSeek-R1-Zero} \cite{guo2025deepseek} have pushed this frontier by training LLMs via reinforcement learning to produce structured reasoning paths, using formatting conventions (e.g., enclosing thoughts in \texttt{<think>} tags) to aid downstream reasoning alignment. While this model demonstrated strong reasoning capabilities, it also exhibited practical limitations, such as decreased readability and occasional mixing of languages.


To mitigate these challenges, \texttt{DeepSeek-R1} incorporated a small quantity of high-quality “cold start” data prior to reinforcement learning (RL). This dataset comprised carefully curated examples, most notably chain-of-thought demonstrations, designed to stabilise early training and improve the coherence of generated outputs. \texttt{DeepSeek-R1} was then trained via a two-stage RL procedure: the first stage targeted improvements in reasoning ability, while the second focused on aligning model outputs with human preferences, thereby enhancing readability and reducing incoherent completions. This multi-phase training strategy enabled \texttt{DeepSeek-R1} to achieve performance on par with OpenAI's \texttt{o1} model across a range of reasoning benchmarks.


While there has been considerable effort in the last two years in developing a large variety of reasoning models, evaluation of such models is still in most cases restricted to a series of widely known mathematics and coding benchmarks, giving the impression to the reader that reasoning for language modelling actually only means solving maths puzzles\footnote{See for instance \url{https://huggingface.co/spaces/opencompass/Open_LMM_Reasoning_Leaderboard}.} \citep{huan2025does}. However, downstream applications of reasoning models can also focus on planning and decision making, where generated reasoning traces could provide an insight into the model's strategy, notwithstanding the pitfalls of being overly reliant on chain-of-thought as an explanation of a model's answers \cite{korbak2025chain,turpin2023language,tanneru2024difficulty,saparov2023language}.

\subsection{Retrieval augmented generation} \label{sec:rag}

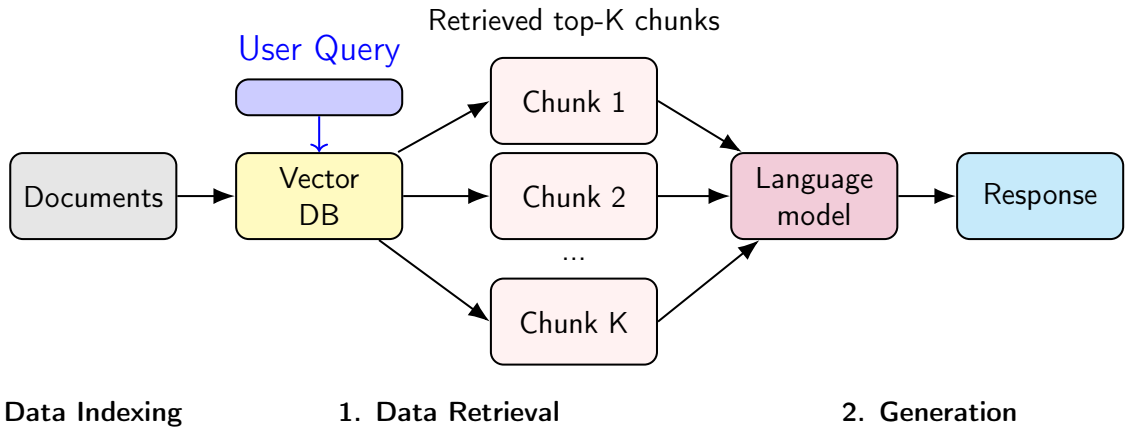
\begin{figure}[t!]
    \centering
    \resizebox{\textwidth}{!}{\begin{tikzpicture}[
  node distance=1.2cm and 1.2cm,
  every node/.style={font=\sffamily},
  box/.style={rounded corners=5pt, draw=black, thick, minimum width=2.3cm, minimum height=1.2cm, align=center},
  box_small/.style={rounded corners=5pt, draw=black, thick, minimum width=2.3cm, minimum height=0.5cm, align=center},
  chunk/.style={rounded corners=5pt, draw=black, thick, fill=pink!20, minimum width=2.3cm, minimum height=1.2cm, align=center},
  arrow/.style={-{Latex[length=3mm]}, thick},
  stage/.style={font=\sffamily\small\bfseries}
]

\node[box, fill=gray!20] (docs) {Documents};
\node[box, fill=yellow!30, right=0.8cm of docs] (vector) {Vector\\DB};
\node[chunk, right=of vector] (chunk2) {Chunk 2};
\node[chunk, above=0.1cm of chunk2] (chunk1) {Chunk 1};
\node[above=0.15cm of chunk1] (topk) {Retrieved top-K chunks};
\node[below=0.1cm of chunk2] (dots) {...};
\node[chunk, below=0.1cm of dots] (chunk3) {Chunk K};
\node[box, fill=purple!20, right=1cm of chunk2] (llm) {Language\\model};
\node[box, fill=cyan!20, right=0.8cm of llm] (response) {Response};

\draw[arrow] (docs) -- (vector);
\draw[arrow] (vector) -- (chunk1.west);
\draw[arrow] (vector) -- (chunk2.west);
\draw[arrow] (vector) -- (chunk3.west);
\draw[arrow] (chunk1.east) -- (llm);
\draw[arrow] (chunk2.east) -- (llm);
\draw[arrow] (chunk3.east) -- (llm);
\draw[arrow] (llm) -- (response);

\node[box_small, fill=blue!20, above=0.5cm of vector] (query_box) {};
\node[above=0.01cm of query_box] (query) {\large\color{blue}User Query};
\draw[->, thick, blue] (query_box) -- (vector.north);

\node[stage, below=2.1cm of docs] {Data Indexing};
\node[stage, below=2.1cm of vector, xshift=1.8cm] {1. Data Retrieval};
\node[stage, below=2.1cm of llm, xshift=1.6cm] {2. Generation};


\end{tikzpicture}}
    \caption{A standard retrieval augmented generation pipeline.}
    \label{fig:rag}
\end{figure}

A \emph{retrieval augmented generation} (RAG) system has two key components (illustrated in Figure~\ref{fig:rag}):
\begin{enumerate}
    \item A \emph{retriever} which retrieves information from some external memory sources. This also involves a pre-processing step to index the knowledge base.
    \item A \emph{generator} (often an LLM) which generates a response based on the retrieved information.
\end{enumerate}

RAG enables LLMs to retrieve relevant document chunks from external knowledge bases, often through \emph{semantic similarity} and \emph{embedding based approaches} \cite{gao2023retrieval}. By utilising an external knowledge base, RAG enables a model to ground its responses in the relevant context, without requiring additional training or fine-tuning, effectively helping it generate relevant  responses and reducing hallucinations \cite{lewis2020retrieval}.

The success of a RAG system heavily depends on the quality of its retriever and its role is to provide the LLM with information from the external database that is most relevant to the query. The retriever has two core functions:
\begin{enumerate}
    \item \emph{Indexing}: pre-processing and chunking the data so that data can be retrieved quickly.
    \item \emph{Querying}: retrieving data relevant to a given query.
\end{enumerate}

Although external data sources may take various forms---including multimodal data (e.g., images, video, audio), tabular datasets, and structured knowledge graphs, this report focuses exclusively on the case where the external memory consists of a corpus of textual documents. In such settings, \emph{document chunking} is typically required to divide each document into smaller, manageable segments that conform to the context window limitations of both the embedding model used in retrieval and the language model used for generation. A common approach is to segment documents based on predefined units such as characters, paragraphs, or token sequences derived from a specific tokenizer. Overlapping chunks are often employed to reduce the risk of splitting semantically important content across boundaries.

To support retrieval in this context, we adopt an embedding-based approach that leverages a \emph{vector store}: a specialised data structure designed for efficient indexing and retrieval of items based on their vector representations, or embeddings. These embeddings are intended to capture the semantic content of the input text and are used to represent the individual document chunks. At query time, the system embeds the input query into the same vector space and performs similarity search against the stored document embeddings to identify the most semantically relevant passages. Common similarity metrics include Euclidean distance and cosine similarity, the latter measuring the cosine of the angle between two vectors and often preferred for its scale-invariance.

More advanced techniques have been developed to improve relevance and contextual alignment between retrieved content and the user query. One such approach is \emph{contextual retrieval} \citep{anthropic2024contextual}, in which short, explanatory contexts are generated and prepended to each document chunk before embedding and indexing, preserving important contextual information that would otherwise be lost when documents are split into smaller pieces. Furthermore, the retrieval process can also be conditioned not only on the query itself but also on additional context, such as preceding dialogue turns or the evolving state of a task \cite{metzler2021rethinking,yu2021few}. This enables the retriever to return passages that are better aligned with the intent and discourse structure of the interaction. Additionally, reranking mechanisms are often employed to refine the initial retrieval outputs \cite{karpukhin2020dense}. These rerankers, typically implemented as lightweight neural models or cross-encoders, score a candidate set of retrieved passages more precisely by jointly considering the query and each document chunk, allowing for improved selection of the most informative context to pass to the generator.

Recent work has also explored retriever-generator co-training, where both components are trained jointly or iteratively in a closed loop \cite{izacarddistilling}. This can lead to tighter coupling between retrieval and generation, with the retriever learning to prioritize passages that the generator can most effectively condition on for producing accurate and coherent responses. Moreover, multi-hop retrieval extends the RAG paradigm by chaining multiple retrieval steps, allowing the system to aggregate evidence from disparate sources across documents \cite{zhang2021answering}. Collectively, these techniques move in the direction of retrieval-aware reasoning, where the retrieval process is optimized not just for relevance, but also for supporting structured inference and faithful generation.

\subsection{Lean language models} 
\label{sec:language_models}

The increasing size of frontier language models has led to a shift in natural language processing (NLP) research, where these models are often accessed as services via dedicated APIs \cite{la2024language,chan2025prompto}. This model-as-a-service approach is sometimes the only available option, particularly for closed models such as those provided by OpenAI. When large-scale open models are released, such as the 405B (405 billion parameters) \texttt{Llama 3.1} \cite{grattafiori2024llama} or the 671B \texttt{DeepSeek-R1} model \cite{guo2025deepseek}, the computational demands for running them, even just for inference, are often prohibitive for many research groups and small organisations. As a result, cloud-based LLM endpoints, such as those offered by Microsoft Azure, provide a practical and accessible way to interact with these models.

However, this approach is not suitable in contexts where data is private or sensitive and must remain on-premises. In such cases, sending data to third-party APIs is not a viable option. This motivates the development and deployment of small-scale language models, designed to operate efficiently in resource-constrained environments while maintaining competitive performance. Techniques such as \emph{quantisation} \citep{jacob2018quantization}, \emph{pruning} \citep{sun2023wanda}, and \emph{knowledge distillation} \citep{hinton2015distilling,gu2024minillm} are commonly employed to reduce model size and computational requirements.

In particular, knowledge distillation, where a smaller ``student'' model is trained to replicate the behaviour of a larger ``teacher'' model, has shown a lot of potential for both reducing the size of an LLM while maintaining overall good performance. For instance, Gemma 2 models \cite{team2024gemma2}, available in sizes ranging from 2 billion to 27 billion parameters, are designed for efficient natural language understanding and generation tasks. While the larger variants are trained from scratch, the smaller variants, such as the 2B and 9B models, utilise knowledge distillation from the larger 27B model to achieve competitive performance. 

These strategies are also adopted when the purpose is to obtain small-scale versions of reasoning models \cite{magister2022teaching}. Knowledge distillation has been used, for instance, by the DeepSeek team \cite{guo2025deepseek} to create distilled versions of their \texttt{DeekSeek-R1} model, adopting Qwen and Llama as starting LLMs and employing 800k samples. An alternative strategy is the one presented by the \texttt{s1} team \cite{muennighoff2025s1}, where the authors fine-tuned 32B Qwen models on only one thousand reasoning traces (initially from Gemini Flash Thinking in \texttt{s1}, and then from DeepSeek in \texttt{s1.1}), highlighting the importance of curating a small set of selected training data versus a large-scale pool of more noisy examples.

These strategies have paved the way for \emph{lean language models}, as a natural counter-point to massive trillion-parameter frontier models. The advantages of small-footprint language models have been widely acknowledged, particularly in the context of agentic systems \cite{wang2024comprehensive,belcak2025small}. The development of novel algorithmic innovations to create highly capable lean models, through strategies such as test-time scaling, is a very active area of research.

Building on top of this previous work, in this technical report we focus on how we can enhance in-domain reasoning capabilities on a small-scale language model, which would be provided with a series of documents to address a user query.

\section{System setup}

In this section we provide an overview of our system covering all its aspects, from the computational infrastructure needed to the pipeline design and frontend interface for chat interactions. 

\subsection{Computational resources}

In this technical report, we present our approach to fine-tuning a
lean reasoning model within a setup that is accessible to relatively
small-scale research labs and industry teams, one that does not depend
on large-scale infrastructure. We followed the configuration described
in \texttt{s1} \cite{muennighoff2025s1} as closely as possible and
used 16 NVIDIA A100 $80\,\text{GB}$ GPUs for the largest models. This
hardware configuration was required to train the 32B model, primarily
due to the large context window employed during training (block size =
$32\,768$ tokens). Such a long context window is needed to make sure the model would be able to rely
on (often long\footnote{Note that one of the key methodological findings
  of \texttt{s1} is the benefit of extending the reasoning process by
  forcing its continuation, instead of allowing the model to end it's generation and output an answer.})
reasoning traces, as well as retrieved documents,
to produce an answer to a user query.

The A100 $80\,\text{GB}$ GPUs offer memory and performance characteristics close enough to the NVIDIA H100s used in the \texttt{s1} study, making them a viable alternative for similar workloads. We conducted experiments on Microsoft Azure and two UK-based academic high-performance computing (HPC) platforms, Baskerville\footnote{\url{https://www.baskerville.ac.uk/}} and Isambard-AI\footnote{\url{https://docs.isambard.ac.uk/}} \cite{mcintoshsmith2024}, evaluating various configurations. Additionally, we leveraged Microsoft Azure's AI Foundry, a suite of tools and APIs designed to streamline integration with foundation models.

\paragraph{Microsoft Azure.} We utilised two virtual machines of type \texttt{Standard ND96amsr A100 v4}, each providing:
\begin{itemize}[noitemsep]
    \item 96 vCPUs and 1800 GB of system memory
    \item 8 NVIDIA A100 80GB GPUs per VM
\end{itemize}

These VMs offered excellent memory utilisation for large-model fine-tuning. We found that using 8 GPUs per node, specifically, two nodes with 8 GPUs each (16 GPUs in total) to fine-tune the 32B model, resulted in more efficient training due to tighter GPU coupling and better memory saturation. In contrast, HPC systems with only 4 GPUs per node required distributed training across more nodes, for example, 6 nodes (24 GPUs in total) on Isambard-AI, which introduced additional overhead and reduced efficiency.

Through \textbf{Azure AI Foundry}, we accessed inference endpoints for the following models:
\begin{itemize}[noitemsep]
    \item OpenAI API models: \texttt{GPT-4o}, \texttt{o3-mini}
    \item DeepSeek models: \texttt{DeepSeek-R1}
\end{itemize}
These endpoints enabled efficient generation of reasoning traces and synthetic user queries as well as final performance testing for frontier LLMs.

\paragraph{Baskerville.} Additional experiments were conducted on Baskerville, a GPU cluster hosted by the University of Birmingham. Each node contains:
\begin{itemize}[noitemsep]
    \item 4 NVIDIA A100 GPUs (40GB or 80GB), connected via NVLink 3.0
    \item High-bandwidth InfiniBand HDR interconnect between nodes
\end{itemize}
We also had access to exploratory nodes with NVIDIA H100 80GB GPUs, which were employed in selected fine-tuning and evaluation runs.

\paragraph{Isambard-AI Phase 1.} We also ran experiments on Isambard-AI, one of the UK's national AI research computing platforms. Phase 1 of Isambard-AI consists of 42 nodes based on the \texttt{aarch64} architecture. Each node includes:
\begin{itemize}[noitemsep]
    \item 4 NVIDIA GH200 Grace Hopper Superchips
    \item Each superchip combines a Grace CPU and a Hopper H100 GPU
    \item Slingshot 11 high-speed interconnect (4 Cassini NICs per node, 200 Gbps each)
\end{itemize}

\paragraph{Compute Usage Summary.} Across the different platforms, we used approximately:
\begin{itemize}[noitemsep]
    \item \textbf{700 GPU hours} on Isambard-AI
    \item \textbf{500 GPU hours} on Baskerville
    \item \textbf{2500 GPU hours} on Microsoft Azure
\end{itemize}
These compute resources enabled comprehensive experimentation across model sizes, fine-tuning strategies, and inference configurations. The final 32B parameter model fine-tuning lasted for approximately 80 GPU hours.

\subsection{Pipeline overview}

Our pipeline comprises multiple steps: indexing a collection using a vector database; retrieving documents via vector similarity to a user query, reasoning about the obtained results and finally generating an answer. In this section, we cover each step of our pipeline, starting from the language model used, which is the central aspect of our system.

\subsubsection{Lean language models}

\begin{figure}[htbp]
  \centering
  \includegraphics[width=\linewidth]{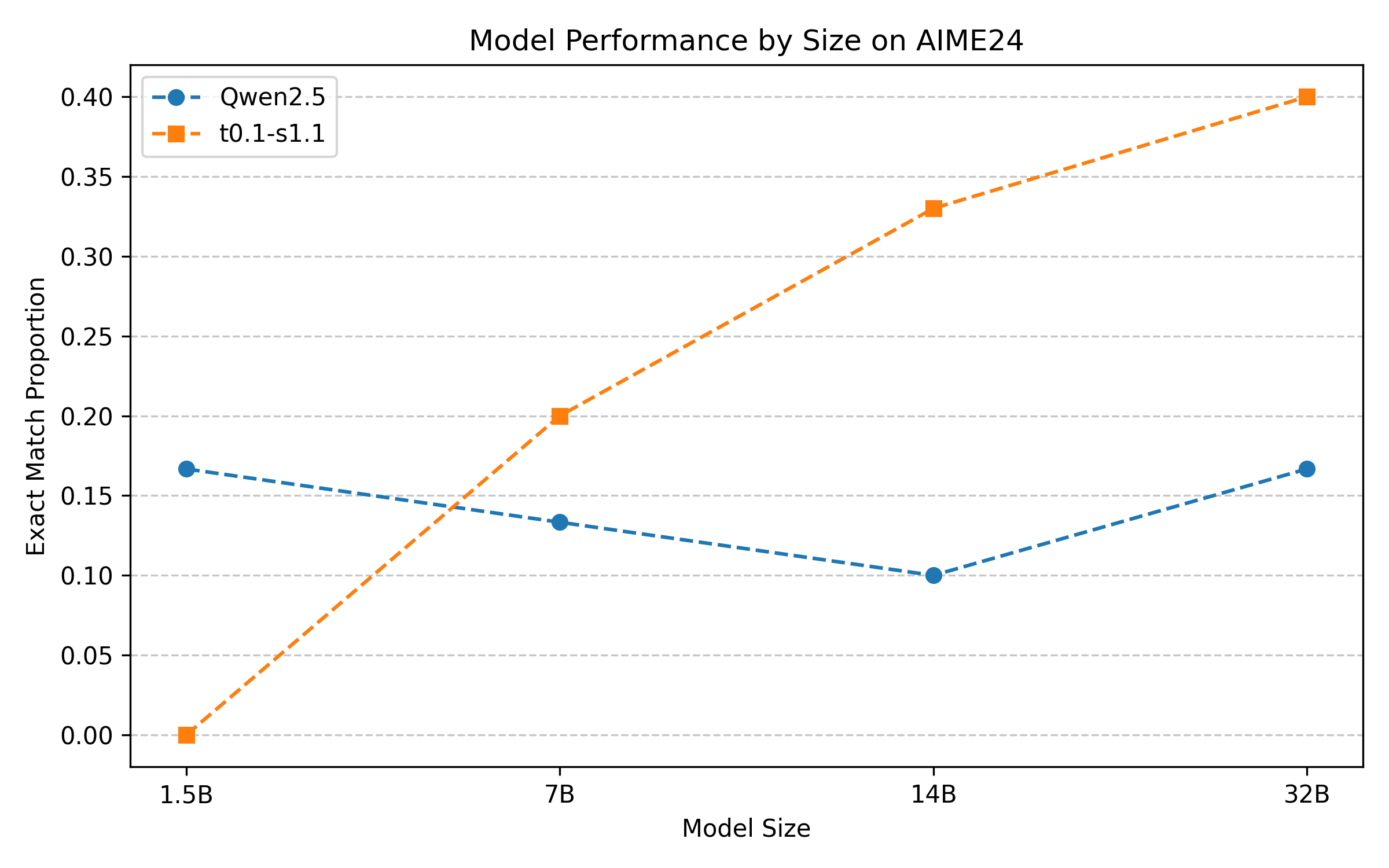}
  \caption{Performance on AIME24 of \texttt{Qwen2.5-Instruct} models and their post-trained versions which are fine-tuned on \texttt{DeepSeek-R1} reasoning traces (reproducing \texttt{s1.1} work, following \cite{muennighoff2025s1}).}
  \label{fig:model-performance}
\end{figure}

Following previous research in test-time scaling \cite{zuo2025ttrl,muennighoff2025s1}, we also employ \texttt{Qwen2.5\-Instruct} models due to their general competitive performance, extended context length and open-source availability. In particular, we focus here on models ranging from 1.5B to 32B parameters, to understand model capabilities at different scales. In Figure \ref{fig:model-performance}, we replicate \texttt{s1.1}\footnote{\texttt{s1.1} (\url{https://huggingface.co/simplescaling/s1.1-32B}) is the successor of \texttt{s1} with better reasoning performance by leveraging reasoning traces from DeepSeek-R1 instead of Gemini. The reasoning traces are available at \url{https://huggingface.co/datasets/simplescaling/s1K-1.1}.} fine-tuning by following the same training procedure as described in \citep{muennighoff2025s1}\footnote{To fine-tune \texttt{Qwen2.5-Instruct} on the \texttt{s1K-1.1} dataset, we followed the training scripts available at \url{https://github.com/simplescaling/s1}.} at different model sizes and examine model performance on the (mathematical reasoning) AIME24\footnote{\url{https://artofproblemsolving.com/wiki/index.php/AIME_Problems_and_Solutions/}} benchmark.

In this setting, the benefit of reasoning capabilities emerge clearly when adopting models with at least 14B parameters, while for smaller models (1.5B), the fine-tuning process ends up negatively impacting model performance. Although the authors also proposed a \emph{budget forcing} approach to control test-time compute by forcefully terminating the model's thinking process or lengthening it by appending ``Wait'' to the model's generation when it tries to end, we did not apply it in this initial experiment in order to be able to compare baseline reasoning capabilities across sizes.

\subsubsection{The retrieval system} \label{sec:retrieval_system}

As described in Section \ref{sec:rag}, a retrieval augmented generation (RAG) system has two key components: a \emph{retriever} and \emph{generator}. For our retriever, we used an embedding based approach where document chunks are indexed and given an vector embedding using a \emph{sentence transformer} model \cite{reimers-2019-sentence-bert}. For querying the vector database, a similarity search is performed to identify the top-$k$ most similar chunks.

As a default embedding model, we adopt \texttt{sentence-transformers/all-mpnet-\allowbreak base-v2}. This model has 109 million parameters and maps sentences and paragraphs to a 768-dimensional vector space. The maximum sequence length of this model is 384 tokens. Note when splitting our documents into chunks, we use a chunk overlap of 50 tokens by default. In practice, we found that this sentence transformer model offered good performance while also being fast and cheap for inference.

We used a \emph{Chroma}\footnote{\url{https://github.com/chroma-core/chroma}} vector database which by default uses an $\ell^2$-norm similarity score metric. In our codebase, we also provide the functionality for the user to use alternative sentence transformer models and an option to use a FAISS \cite{douze2024faiss} database instead. In our use case, we found that Chroma and FAISS offered similar retrieval performance, but Chroma was slightly faster.

Our retrieval system uses \emph{full document retrieval} whereby if any chunk from a particular document is found within this top-$k$ set of chunks, our system proceeds to retrieve the entire original document from which that chunk originated. This ensures that the LLM receives the complete context surrounding the relevant information, even if only a small portion of the document was initially flagged as relevant.

\subsubsection{Synthetic data generation}

Given a collection of documents, we use a language model to
generate a series of queries that are: (a)~relevant for a selected
document from the collection; and (b)~rely on understanding
information in the content of the document to provide a valid
answer. In this way we are able to generate a large set of user
requests and know in advance the correct answer (i.e., specific
document and specific piece of information in the document). In order
to build a challenging evaluation dataset, we also prompted the model
to produce more complex queries, for instance vague user
requests. More details about this are given in the case-study
described in the experimental section.

In our experiments, we have relied on OpenAI's \texttt{GPT-4o} in order to generate a set of high quality queries for testing the usefulness of our system, but our codebase allows users to customise their own set up, e.g., choice of model, prompt template and number of queries to generate. This way users could employ a local LLM for this step of the process too.

\subsubsection{Reasoning traces}

Given a synthetically generated query and a set of retrieved documents from our collection, we prompt a large reasoning model to obtain its reasoning traces and final answer. For this step, we use \texttt{DeepSeek-R1} in our experiments, but the user can easily choose a different reasoning model. Through this process, we generate a dataset of reasoning traces, each comprising a query, a set of retrieved documents, the reasoning process and the model's final answer.

\subsubsection{Fine-tuning}

We use the reasoning traces to fine-tune a smaller model in order to to enhance its capabilities at test-time. The goal is that the model should start producing a ``reasoning process'', similar to the one of \texttt{DeepSeek-R1}, before providing its final answer, and that such reasoning process should improve overall performance.

We follow the approach described in \texttt{s1} \cite{muennighoff2025s1} and perform supervised fine-tuning on next token prediction of \texttt{Qwen2.5-Instruct} models (ranging from 1.5 to 32B parameters), with basic hyperparameters. The main challenge, which distinguishes our work from the set up of \texttt{s1}, is the fact that each of our model responses are much longer compared to the ones in the \texttt{s1} dataset, as they include reasoning traces \emph{and} a set of retrieved documents. This is due to the fact that we retrieve full documents rather than chunks as described in Section \ref{sec:retrieval_system}. In particular, in our first attempt to create reasoning traces with setting the number of retrieved documents to $5$, the average token length of the training examples using the Qwen2.5 tokenizer was $74\,641$. By comparison, the \texttt{s1K}\footnote{\url{https://huggingface.co/datasets/simplescaling/s1K_tokenized}} and \texttt{s1K-1.1}\footnote{\url{https://huggingface.co/datasets/simplescaling/s1K-1.1_tokenized}} datasets have an average token length of $9\,109$ and $26\,969$, respectively.

In order to train our model maintaining the same computational resources, we have employed automatic document summarisation, to reduce the length of the input context, while still being able to benefit from the retrieved materials. 

\subsubsection{Retrieved document summarisation}

To make the training process possible, we reduced the length of the context by tackling in particular the size of the original documents in our collection, while maintaining their core information. We employed \texttt{Qwen2.5-32B-Instruct} to generate a summarised version of each document in our collection, reducing the size of each document by 85\% of its original length. In our experiments, we also make sure that this has no impact on retrieval performance, but users should assess that this is consistent in other applications. By using summarised documents, the average token length of our reasoning traces was reduced to $7\,544$.

An alternative approach to basic document summarisation would be query-aware document summarisation, namely to summarise each document \emph{on the fly}, after having it retrieved. This way the summarisation process would know which part to keep (the ones relevant to the user query) and which to exclude from the summary. Note however that this would slow down the system, as it would be an additional LLM action for each user query, compared to having a static, summarised version of the collection, as in our case.

\subsection{Conversational interface}

In the next parts, we describe how we connect all these elements to allow a fluid multi-turn interaction between the system and a user. 

\subsubsection{Orchestration for chat interactions} 
\label{sec:rag_conv}

To bring all these components together, we used the \emph{LangChain}\footnote{\url{https://www.langchain.com/}} framework in Python to develop our RAG pipeline and combine this with our fine-tuned language model to create a conversational chat bot application. 

For our RAG application, we wanted to allow the user to have a back-and-forth conversation whereby the language model is given the previous conversation history and the retrieved context to construct a response. To incorporate historical messages, it is necessary to use the conversation history and a \emph{prompt template}. Prompt templates turn raw user/human-AI chat interactions into a format that the language model can work with and generate a response for. Typically chat templates are language model-specific, meaning different language model families such as Llama \cite{grattafiori2024llama}, Gemma \cite{team2024gemma, team2024gemma2} and Qwen \cite{bai2023qwen, qwen3technicalreport} use different chat templates. For example, for \texttt{Qwen2.5-Instruct} models use the following format to indicate the role and content of a given interaction:
\begin{verbatim}
<|im_start|>{role}
{content}<|im_end|>
\end{verbatim}

The \texttt{role} can be one of \texttt{user}, \texttt{assistant} or \texttt{system}. A \texttt{system} message can be useful to instruct the model to perform certain actions or undertake different characteristics such as the tone or style to use. Given a list of user-AI chat interactions, we can use the Qwen prompt template to construct a prompt to the language model such as:
\begin{verbatim}
<|im_start|>system
You are a helpful assistant.<|im_end|>
<|im_start|>user
hello<|im_end|>
<|im_start|>assistant
Hello! How can I assist you today?<|im_end|>
\end{verbatim}

For presenting retrieved context from a knowledge base to the model, we can construct a system prompt template (see Appendix \ref{app:rag_conv_template} for the system prompt we used for the exemplary application described in Section \ref{sec:nhs_application}) which defines the task for the model and presents the retrieved context as well as additional information such as the demographics of the user.

Note that we could alternatively have used a user prompt template whereby the retrieved context is presented in the user message. We chose not to do this to limit the growth of the conversation history context length, since presenting retrieved context in the user message leads to the retrieved context staying in the conversation history as the chat develops. In our case, when retrieval is used, the retrieved context is provided to the model in the system prompt and so it can be different throughout the conversation.

The language models used here all have finite context windows. Consequently, as conversations accumulate long message histories, it might be necessary to reduce the size of the chat history. We do this by trimming the history based on token count. Note that we never delete the system prompt from the history and only remove the oldest chat interactions if necessary. 

\subsubsection{Retrieval as a tool}

In a standard RAG setting, we may simply use the last user message as the query to the retriever. However, there are two key issues with this simple approach. 

Firstly, in many conversation interactions, the user message itself is not informative enough on its own to be a useful query for the retriever. Common situations where this is the case are follow-up questions based on the previous conversation history. For instance, consider the following conversation:
\begin{quote}
    \emph{User}: I have been having headaches recently, what are common ways to alleviate a headache?
    
    \emph{AI}: Common ways to alleviate a tension headache include taking over-the-counter pain relievers like ibuprofen or acetaminophen, applying a warm or cool compress to your head or neck, and practising relaxation techniques.
    
    \emph{User}: Where can I buy them?
\end{quote}

In a standard RAG setting, the query ``Where can I buy them?'' is ambiguous without the context of the full conversation. 

Secondly, often user messages may be simple enough that the model would not require retrieval. For instance, for simple user messages such as ``Hello'', it would be cheaper to avoid retrieval and for the model to respond directly.

To address these two issues, it is possible to treat retrieval as a \emph{tool} that a model has access to. For our purposes, a \emph{tool} is an association between a function and its schema that defines the function's name, description and arguments. This schema is then passed to the language model and the language model can decide to use that tool by defining the name of the tool to use and the arguments to use. This approach leverage \emph{tool-calling}\footnote{For more details on tool-calling in LangChain, see \url{https://python.langchain.com/docs/concepts/tool_calling/}.} (sometimes called \emph{function-calling}) which is now commonly supported in many modern chat models and endpoint providers.

In such a setting, we refer to the language model as an \emph{agent} \cite{yao2023react,shinn2023reflexion} which combines language generation with actions. Generally, an \emph{agent} refers to anything that can perceive its environment and act upon that environment \cite{russell:1995:AI}. The set of actions that an agent can perform is defined by the tools it has access to. For our RAG pipeline, the language model can be considered an agent and the tool is the text retriever. Therefore, for a given user message, the model can decide to either query the retriever or to respond directly in natural language. Note that in our use-case, the language model ideally only decides to not to use retrieval in simple user messages since we want most model responses to be grounded with data from the knowledge base. 

In the case that the language model decides to use the retriever, a tool-call is made and the language model decides the query to use. Since the language model has access to the conversation history, it can leverage previous chat interactions to choose a relevant query to the vector database. 

\begin{figure}[ht]
    \centering
\resizebox{\textwidth}{!}{\begin{tikzpicture}[
  node distance=1.2cm and 1.2cm,
  every node/.style={font=\sffamily},
  box/.style={rounded corners=5pt, draw=black, thick, minimum width=2.3cm, minimum height=1.2cm, align=center},
  box_small/.style={rounded corners=5pt, draw=black, thick, minimum width=2.3cm, minimum height=0.5cm, align=center},
  box_empty/.style={rounded corners=5pt, minimum width=2.3cm, minimum height=0.5cm, align=center},
  chunk/.style={rounded corners=5pt, draw=black, thick, fill=pink!20, minimum width=2.3cm, minimum height=1.2cm, align=center},
  arrow/.style={-{Latex[length=3mm]}, thick},
  arrow_dash/.style={-{Latex[length=3mm]}, dashed},
  stage/.style={font=\sffamily\small\bfseries}
]

\node[box_small, fill=blue!20] (query_box) {};
\node[above=0.01cm of query_box] (user_query) {\large\color{blue}User Query};

\node[box, fill=gray!20, right=0.8cm of query_box] (conv_agent) {Conversational\\agent};
\node[box_empty, right=0.8cm of conv_agent] (dots) {};
\node[box_empty, right=of dots] (dots2) {};
\node[box, fill=yellow!30, above=1.4cm of dots] (retriever) {Retriever};
\node[box, fill=purple!20, above=1.4cm of dots2] (llm) {Reasoning\\model};
\node[box, fill=cyan!20, right=0.6cm of dots2] (response) {Response};

\draw[arrow] (query_box) -- (conv_agent);
\draw[arrow_dash] (conv_agent) -- (retriever);
\draw[arrow_dash] (conv_agent) -- (response);
\draw[arrow] (retriever) -- (llm);
\draw[arrow] (llm) -- (response);

\node[stage, above=0.8 cm of conv_agent, xshift=0.4cm] {Generate query};
\node[stage, above=0.4 cm of conv_agent, xshift=0.4cm] {to retriever};
\node[stage, below=0.1cm of dots, xshift=1.6cm] {Generate a response directly};


\end{tikzpicture}}
    \caption{A conversational retrieval augmented generation
      pipeline.}
    \label{fig:rag-conv}
\end{figure}
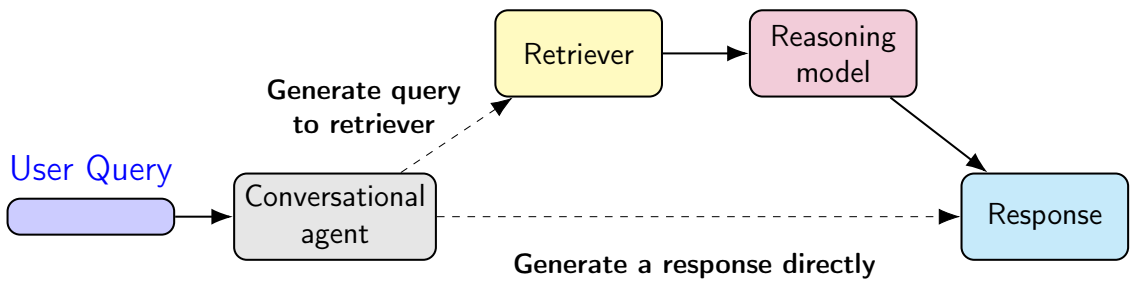

Figure \ref{fig:rag-conv} presents the full flow of a given query. First the query is presented to a \emph{conversational agent language model} which decides whether to query the retriever, or to respond directly (in the case of simple messages). The system prompt for this conversational agent is presented in Appendix \ref{app:conv_template}. If the conversational agent decides to perform a tool call to the retriever, it writes a query to the retriever given the chat history and retrieves relevant document chunks from the knowledge base. Lastly, we present the model the retrieved context through the system prompt of the reasoning model as described above to generate a response given the chat history. Note that there is flexibility in the choice of language models at the different stages as the conversational agent language model which decides whether retrieve first and respond or respond directly can be different to the language model that uses the retrieved context to generate a response. In our final model, we used \texttt{Qwen2.5-Instruct-32B} as the conversational agent language model and \texttt{t0-1.1-k5-32B} as the RAG language model, since we found \texttt{Qwen2.5-Instruct-32B} to be more consistent for tool-calling.

\subsubsection{Frontend interface}

The final piece of code in this pipeline is the \emph{frontend}: a static web application written in the Svelte framework,\footnote{\url{https://svelte.dev/}} which allows users to converse with the language model in an interface largely resembling that of OpenAI's ChatGPT website.
The website is deployed on GitHub Pages,\footnote{\url{https://alan-turing-institute.github.io/t0-1}} and interacts with the Python chatbot (the \emph{backend}) via a REST API.
Users can create multiple separate conversation threads, each with unique identifiers, and switch between them at will.
The responses from the chatbot are parsed by the browser: by default, only the main answer is shown in full to the user, with reasoning traces available via a dropdown toggle.

In general, the backend has sole responsibility for storing the conversation history: the frontend only serves as a way to parse and display this information to an end-user.
Thus, the state of the frontend is fully derived from that of the backend; this design prevents potential inconsistencies that may arise if the frontend were to store its own copy of the conversation history.

Because the Python chatbot is exposed over HTTP, and modern browsers do not (by default) allow HTTPS pages to make requests to HTTP endpoints, it was further necessary to set up an Nginx reverse proxy to act as an intermediary.
Thus, insecure HTTP connections were handled by the proxy, and the website would only ever see connections to a secure HTTPS URL.
This could, if desired, be replaced with Caddy.\footnote{\url{https://caddyserver.com/}}

For this proof of concept, it was considered unnecessary to introduce authentication such that each conversation could be uniquely associated with a user.
Thus, every visitor to the webpage can see every conversation available.
Implementing authentication using OAuth2 would be an obvious next step for a more serious deployment.

\section{Exemplary Application} \label{sec:nhs_application}

\begin{figure}[t]
  \centering
  \includegraphics[width=\linewidth]{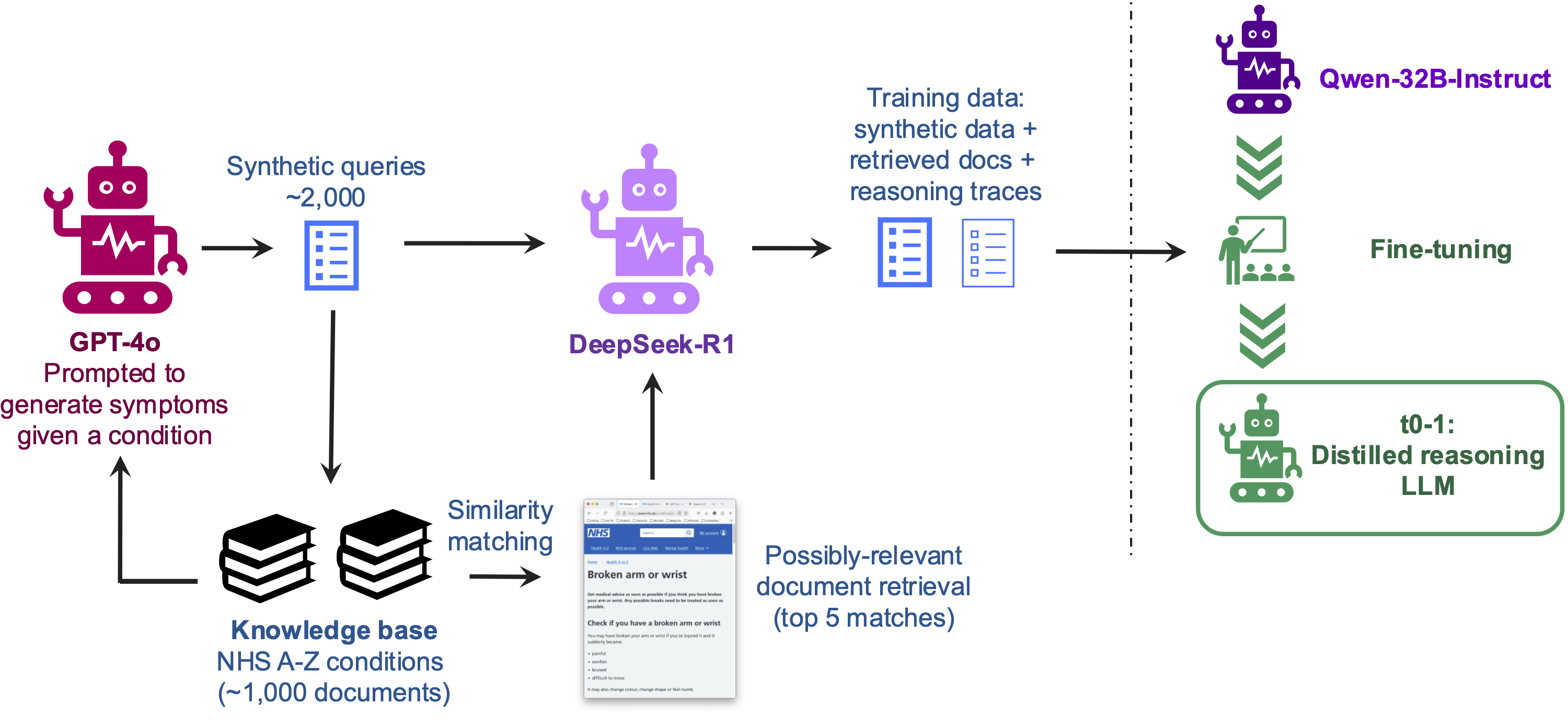} 
  \caption{Overview of our pipeline, covering the central aspects of the process: synthetic data creation, information retrieval, reasoning trace generation, and model fine-tuning.}
  \label{fig:pipeline_overview}
\end{figure}

Research in test-time scaling and reasoning generally focuses its application in the mathematics and code domains \cite{zuo2025ttrl,guo2025deepseek,muennighoff2025s1}, since addressing such questions might require a reasoning process, assessing the correctness of a final answers is straightforward and there are many datasets available \cite{zhang2025survey}. 

In our experiments, we instead focus on a different scenario, which hopefully is closer to real applications which aim to leverage upon the combination of information retrieval and reasoning for decision making. We consider the body of knowledge provided by the NHS A-to-Z condition website.\footnote{\url{https://www.nhs.uk/conditions/}} For each of the almost $1\,000$ conditions listed, a webpage provides information about it and a series of possible next actions, depending on the patient symptoms (for instance asking for an urgent GP appointment or going directly to A\&E). We consider this an interesting setting for testing our model, as it would require a retrieval component (to interpret the user query and search across the available documents) and a reasoning element (to interpret the patient symptoms and decide for the best next step, while staying grounded on the provided information). An overview of the process, which we will cover step by step in this section, is shown in Fig.~\ref{fig:pipeline_overview}.

It is important to emphasize that the prototype presented here is not intended as a tool for providing medical advice. Rather, the medical domain is used purely as a demonstration of its potential applicability across sectors that rely on private, specialized knowledge and complex queries to support decision-making.

\subsection{Dataset} \label{sec:nhs_dataset}

We collected conditions by scraping all webpages under the NHS Conditions subdomain, obtaining 990 different conditions.\footnote{The script for downloading the pages is available here: \url{https://github.com/alan-turing-institute/t0-1/blob/main/scripts/Makefile}.} We then removed the condition ``Mental Health'', as the page is actually a collection of conditions structured in a different way compared to the rest of the collection. The remaining $989$ are organised in a single dataset (as a JSON Lines file) containing the name of the condition (the page title), the entire content of the page (if the condition webpage contained multiple sub-pages, we concatenated the content in a single stream of text) and a summarised version of the content of the page obtained using \texttt{Qwen2.5-32B-Instruct}. The prompt used for this task can be found in Appendix \ref{app:summarisation_prompt}.

With this prompt, we reduced the size of each document by 85\% of its original length, while maintaining core information relevant for the task, as every page contained boiler-plate text and repeated information. However, it is true that reducing the content size so drastically might lead to retrieval issues in the downstream task, so we have compared retrieval performance when operating on the full content or only on the summarised versions (see Table \ref{tbl:embedding_results}).

\subsection{Synthetic user queries} \label{sec:eval_synthetic}

Given the full content of a condition page and a predetermined disposition (self-care, urgent primary care and A\&E) we prompted \texttt{GPT-4o} to generate a synthetically-generated patient query (or to refuse in case our request was not applicable, e.g., the disposition was not in line with the possible outcomes on the page content). We also ask the model to generate general patient information (e.g., age, occupation and social support), in line with the condition and disposition. We control the patient sex, as in early experiments we noticed the model was over-generating examples of female patients over male ones. In order to better understand how our approach would fail with more complex requests, we ask the model to generate three types of queries:

\begin{itemize}
    \item \textbf{basic}: Based on a single condition page, the query mentions relevant symptoms.
    \item \textbf{hypochondriac}: Based on a single condition page, the query mentions relevant symptoms plus other unrelated complaints and expressions of excessive anxiety.
    \item \textbf{downplay}: Based on a single condition page, the query downplays the severity of the symptoms.
\end{itemize}

While basic represents the most common type of requests for this type of system, hypochondriac and downplay challenge the pipeline by offering either too much or too little information about the condition and severity level.

The prompt used for generating the synthetic data is available in Appendix \ref{app:synthetic_data_prompt}. As an example, the following synthetic request was generated with an input prompt for a basic query from a woman which should be matched with the condition hip-replacement and the disposition should be urgent primary care. The rest of the content in the example is generated by \texttt{GPT-4o}:

\begin{tcolorbox}[colback=gray!10, colframe=black, title=Example]
\textbf{General patient information:} age: 65, sex: female, occupation: retired Teacher, social support: I live with my husband who helps me around the house., medical history: I have osteoarthritis and occasionally take over-the-counter pain relief. No other significant conditions.\newline

\textbf{Symptoms Description:}
I had a hip replacement about two weeks ago, and initially everything seemed fine, but now I'm noticing some worrisome symptoms. The area around my hip is swollen and red, and it's feeling more tender than it did before. I'm also a bit shivery, and when I checked, I had a temperature of about 38.5C this morning. I feel slightly more pain in my leg when I try to walk. I don't see any pus from the wound, but I'm worried it might be the start of an infection. Should I get it checked urgently?
\end{tcolorbox}

Using the process described above, we generated two datasets, one
containing $1\,000$ synthetic queries as an evaluation set and a
second one containing $2\,000$ synthetic queries as a dataset to be
used for fine-tuning. To ensure no overlap between our evaluation and fine-tuning datasets, we identified queries which had the same combination of condition and disposition. We then removed these queries from the fine-tuning dataset and generated further data to bring our total number of queries back up to $2\,000$. A clinician reviewed a subset of our
synthetically generated requests, confirming their suitability and
different levels of complexity. We relied on a frontier LLM (\texttt{GPT-4o}) for this step; this part of our approach would
require a different strategy if adopted on private data held in secure
environments. As alternatives, we suggest the following options: if
possible, obtaining examples of real queries on the body of
knowledge under study, which would then perfectly reflect the type of
process to automatize. Alternatively, we recommend adopting a small-scale model
that can be run on-premise like \texttt{Qwen2.5-32B-Instruct} and then
following the rest of our approach for synthetic data generation. In this case, careful evaluation of the synthetic data quality is essential to ensure its effectiveness for downstream tasks. 

\subsection{Retrieval performance}

Using the evaluation set of $1\,000$ synthetic user queries, in
Table \ref{tbl:embedding_results} we report the performance of our
retrieval component, when indexing full condition pages or the
summarised versions. As described in the overview of the system setup,
when indexing full documents, these are divided into chunks to allow
the identification of specific passages in the condition page that are
relevant to the query. Chunking increases the number of documents in
the database from $988$ to $5\,824$. We consider a series of cutoffs
$k$ when testing the retrieval system, ranging from $k=1$, where only
the most similar document in the vector database is returned, to
$k=100$.

The metric $p@k$ refers to the fraction, $p$, of queries for which
the correct condition is present among the $k$ returned
documents. For example, $p@5$ is the proportion of queries for which
the correct condition appears among the five most similar documents
retrieved. Note that when retrieving summaries, the cutoff number
corresponds to the number of condition pages since the summarised documents all had context length smaller than that of our embedding model ($384$), while for full documents
it corresponds to the number of \emph{chunks} returned.
 
Knowing the performance at different cutoffs allows us to choose a
suitable number of retrieved documents to use when combining retrieval
and reasoning. A higher number of documents will guarantee, in most
cases, that the correct condition is retrieved, but will also lead to
a longer context for the downstream LLM, impacting fine-tuning and
reasoning.

\begin{table}[ht]
  \centering
  \caption{Retrieval accuracy at different cutoffs when indexing full pages (which are then divided into \emph{chunks}) versus summaries of the same pages.}\label{tbl:embedding_results}
  \begin{tabular}{@{}lcrrrrrr@{}}
    \toprule
    \textbf{Input} & \textbf{Documents} & $p@1$ & $p@5$ & $p@10$ & $p@30$ & $p@50$ & $p@100$ \\
    \midrule
    Full pages & $5\,824$ & 0.47 & 0.68 & 0.78 & 0.87 & 0.91 & 0.94 \\

    Summaries & 989 & 0.51 & 0.76 & 0.83 & 0.93 & 0.96 & 0.98 \\
    \bottomrule
  \end{tabular}
\end{table} 

Given the performance shown in Table \ref{tbl:embedding_results}, in
our following experiments we consider indexing summaries of documents,
as this consistently leads to higher retrieval performance. Note that
for our chosen use-case, we know what type of queries the system will
receive, so we have summarised content in a way that retains that
information. For a more general retrieval system not only focused on
determining condition and disposition, it may be more advisable to
index full pages. Regarding the cutoffs, we initially experimented with $k=5$
and $k=30$, which retrieved the
correct condition in respectively $76\,\%$ and $93\,\%$
of queries. Regarding the fine-tuning process and final experiments detailed below, we use $k=5$ retrieved documents, as this results in a manageable context length given our resource constraints, and is easier to explore via the user interface. Note, however, that this choice imposes a maximum achievable prediction accuracy of 76\%, since the reasoner only considers the retrieved documents as potentially relevant. To address this limitation while maintaining a short list of retrieved documents, we explored several strategies, including reranking the results via an additional LLM call (similar to \cite{carraro2024enhancing}) and rephrasing user queries, inspired by the reasoning-and-acting approach of the ReAct framework \cite{yao2023react}. However, these alternatives did not consistently lead to significant improvements in our setting.

\subsection{Fine-tuning process} \label{sec:fine_tuning_process}

For each synthetic query in our $2\,000$ example dataset, we prompt
\texttt{DeepSeek-R1} with the summarised content describing the
potentially relevant conditions using $k=5$ as retrieval cutoff. The
prompt template used to generate the reasoning traces can be found in
Appendix \ref{app:reasoning_traces_prompt}. Through this process,
each query is then structured such that it has:
\begin{itemize}
    \item Five retrieved (summarised) conditions pages;
    \item the reasoning process from \texttt{DeepSeek-R1}, to determine based on them the correct condition and disposition;
    \item the final provided answer by the model.
\end{itemize}

These components are all concatenated into a single stream of text for each query, in order to then fine-tune a series of small versions of \texttt{Qwen2.5-Instruct} models for next token prediction. What we expect to see is that the model would start producing a \emph{thinking} or \emph{reasoning} process at test-time focused on the content of the retrieved documents in relation to the user query, before generating the answer. Such process should enhance its capabilities versus a non-reasoning baseline or a general purpose reasoning model (e.g., \texttt{s1}).

The fine-tuning parameters were selected based on recommendations from \cite{muennighoff2025s1}. The most important configuration choices are as follows:
\begin{itemize}
    \item \textbf{Epochs}: $5$
    \item \textbf{Learning Rate}: $10^{-5}$ with cosine scheduler
    \item \textbf{Batch Size}: $1$ (per device)
    \item \textbf{Precision}: \texttt{bfloat16 (bf16)}
    \item \textbf{Block Size}: $32\,768$
    \item \textbf{Sharding}: FSDP (\texttt{full\_shard auto\_wrap})
    \item \textbf{Gradient Checkpointing}: Enabled
    \item \textbf{Optimizer}: Adam with weight decay of $10^{-4}$, $\beta_1 = 0.9$, $\beta_2 = 0.95$
    \item \textbf{Evaluation Frequency}: Every $50$ steps
\end{itemize}

All models were trained using the same fine-tuning parameters as described above. GPU and system configurations varied due to availability and cost considerations. The training setups were as follows:
\begin{itemize}
    \item \textbf{1.5B, 3B, 7B models}: $4\times \text{A100}$ $80\,\text{GB}$ GPUs on Baskerville (1 node)
    \item \textbf{14B model}: $16\times \text{A100}$ $80\,\text{GB}$ GPUs on Baskerville (4 nodes)
    \item \textbf{32B model}: $16\times \text{A100}$ $80\,\text{GB}$ GPUs on Azure (2 VMs)
\end{itemize}

\subsection{Condition and next action prediction} \label{sec:cond_next_action_pred}

In this section, we measure whether retrieval augmented reasoning
improves the performance of a lean language model. In
Table \ref{tbl:llm_accuracy}, we report the performance of our 32B
parameter fine-tuned model (named \texttt{t0-1.1-k5-32B}) on two tasks: (i)~determining the condition of a synthetic patient, given the textual
description of the symptoms; and (ii)~establishing the next course of
actions among the options suggested in the document content. We assume
that a baseline non-reasoning model (\texttt{Qwen2.5-32B-Instruct})
would already have some general knowledge to be able to
perform such task at a decent level. However, the integration of
retrieval and reasoning capabilities should improve performance, as in
domain evidence would be additionally offered to the model. The prompt
templates used to evaluate model performance on predicting the
condition and the suitable next action are given in Appendix
\ref{app:cond_next_action_pred_prompt}. To provide a comparison with
other systems, we report performance of two recent comparable lean
reasoning models, \texttt{s1.1-32B} \cite{muennighoff2025s1} and
\texttt{Qwen3-32B} \cite{qwen3technicalreport} and also a series of
state-of-the-art large language models (\texttt{GPT-4o},
\texttt{o3-mini}, \texttt{DeepSeek-R1}), to understand overall
frontier performance. For \texttt{t0-1.1-k5-32B} and \texttt{s1.1-32B} we use budget-forcing to control test-time compute as described in \citep{muennighoff2025s1}.\footnote{In particular, we enforce a maximum token count of $1024$ and suppress the generation of the end-of-thinking token delimiter a maximum of $3$ times.} 
\begin{table}[ht!]
  \centering
  \caption{Condition and disposition accuracy by LLM and $k$ value. We report first the examined
    lean language models and then a series of frontier large language
    models. A dash (--) indicates a $k$ value of zero: that is, the model was provided with no retrieved context, as a baseline. Note that for all models relying on a retrieval component, their maximum accuracy achievable for condition is 0.76, as already discussed regarding Table \ref{tbl:embedding_results}. All reported values are average accuracies over 10 runs. Standard deviations were consistently around 0.01 and are omitted for clarity.}\label{tbl:llm_accuracy}
  \begin{tabular}{@{}lcrr@{}}
    \toprule
    \textbf{LLM} & $k$ & \textbf{Condition} & \textbf{Disposition}
    \\
    \midrule
    \multicolumn{4}{@{}l@{}}{\textbf{Lean Language Models}} \\[1ex]
    Qwen2.5-32B-Instruct & -- & 0.38 & 0.46 
    \\
                 & 5 & 0.54 & 0.50 
    \\
    \\[1ex]

    t0-1.1-k5-32B  
                 & 5 & \textbf{0.56} & \textbf{0.51} 
    \\ 
    \\[1ex]


    s1.1-32B 
                 & 5 & 0.49 & 0.46  
    \\
    \\[1ex]
    Qwen3-32B 
                 & 5 & 0.53 & 0.48 
    \\
    \midrule
    \multicolumn{4}{@{}l@{}}{\textbf{Frontier Language Models}} \\[1ex]
    
    GPT-4o & -- & 0.49 & \textbf{0.56} 
    \\
                 & 5 & 0.56 & 0.54  
    \\
    \\[1ex]


    o3-mini & -- & 0.27 & 0.54 
    \\
                 & 5 & \textbf{0.57} & \textbf{0.56}  
    \\
    \\[1ex]
 

    DeepSeek-R1 & -- & 0.44 & 0.53  
    \\
                 & 5 & 0.56 & 0.51  
    \\
    \bottomrule
  \end{tabular}
\end{table}

As presented in Table \ref{tbl:llm_accuracy}, a 32B parameter model already has some core knowledge in the topic at hand, as it is able to identify the correct condition from the synthetic patient description in $38\,\%$ of the cases and predicts the correct next course of action $46\,\%$ of the time. This starting point assessment is necessary to establish the initial capacity of the adopted language model. Depending on the considered task and domain of knowledge, performance will vary, especially when applications are focused on a specific body of knowledge that is not widely available, such as materials only shared on the intranet of an organisation. For comparison, a frontier non-reasoning model, such as \texttt{GPT-4o}, starts with an accuracy of $49\,\%$ for conditions and $56\,\%$ for disposition.

Providing retrieved documents helps models in better predicting for the two tasks, with a performance increase of over $15\,\%$ for condition accuracy for the relatively small \texttt{Qwen2.5-32B} and $7\,\%$ for \texttt{GPT-4o}. Qwen also gains a 4 percentage points improvement for disposition accuracy, while instead \texttt{GPT-4o} sees its performance slightly decreasing, potentially due to more ambiguity from the amount of information now available. Large performance improvements for condition accuracy are also observed for the two frontier reasoning models examined (\texttt{o3-mini} and \texttt{DeepSeek-R1}), highlighting the substantial benefit of incorporating a retrieval component to provide relevant in-domain evidence to the model.

Moving to the lean models examined, we can see the additional benefit of integrating a thinking process before providing the final answer. Our \texttt{t0-1.1-k5-32B}, which has been fine-tuned on in-domain reasoning examples, further increases the accuracy in determining the correct condition, with an almost 20\% performance improvement compared to the base Qwen model and an additional small improvement against using retrieval alone. The results for condition accuracy outperform those of other leaner reasoning solutions examined (\texttt{s1.1-32B} and the newly released \texttt{Qwen3-32B}), bringing the model into the same realm as massively larger frontier reasoning models such as \texttt{o3-mini} and \texttt{DeepSeek-R1}.

We do not see a similar drastic improvement for determining the disposition. While the model performs better than the base \texttt{Qwen2.5-Instruct-32B} and of general purpose lean reasoning models, it does not reach the performance of a frontier model as \texttt{o3-mini}. We believe this is due to the reasoning traces provided by \texttt{DeekSeek-R1}, which set the limit of expertise that can be distilled into smaller models, compared to the better reasoning process of \texttt{o3-mini} for dispositions. It is also important to note that \texttt{o3-mini} shows very low performance for condition accuracy when prompted without retrieved documents. From inspecting the outputs, it seems that \texttt{o3-mini} loses focus when making a prediction without supporting evidence, and ends up considering too many possible scenarios.

The main outcome of this evaluation is that a retrieval augmented lean reasoner model can shortlist the correct condition among a few options in 76\% of cases (as shown in Table \ref{tbl:embedding_results}), and predict the correct condition in 56\% of cases, comparable to state-of-the-art frontier models (as shown in Table \ref{tbl:llm_accuracy}). This serves as a strong starting point before initiating a conversation with the user, which would allow the model to gather additional information, expand its search in the database, and more accurately narrow down the possible conditions and patient disposition (see an example of the conversational interface in Figure \ref{fig:interface}).

\section{Discussion}

In this section we present three aspects that should help orienting future implementations based on our technical report. We start by covering the trade-offs between general purpose versus domain specific reasoning, then we examine how to further reduce model size and we conclude with an overview of our system's frontend as a prototype directly available in our GitHub repository.

\subsection{General vs domain-specific reasoner}

The major infrastructural complexity presented in our technical work is fine-tuning a 32B parameter model for enhancing its in-domain reasoning process using reasoning traces from a larger frontier model. Alternatively, we could have adopted a more general purpose reasoning model of the same size, such as \texttt{s1.1}. In Table \ref{tbl:llm_accuracy} we have highlighted the overall better performance of our approach and in this section we dig deeper in the comparison by considering the variety of queries tested.

In Table \ref{tbl:error_analysis}, we see that on basic type of queries, our model accurately identifies the correct condition in 52\% of the cases and the correct disposition in 64\%, even before starting a conversation with the user. For determining the right condition, performance increases to 62\% for hypochondriac queries, as many more additional details are added, however both in this case and for what concerns downplay queries performance for disposition drops to around 45\%, exactly because they have been generated to challenge the model. Compared to our model, which has been fine-tuned for in-domain reasoning, a general-purpose reasoning model such as \texttt{s1.1} performs 3 to 12\% worse, depending on the query type and evaluation metric. Furthermore, when analysing the types of errors made, we observe over a 40\% increase in underestimation errors---such as predicting ``Urgent Primary Care'' instead of the correct ``A\&E''---when using \texttt{s1.1} instead of \texttt{t0-1.1-k5-32B} on basic queries.
 
\begin{table}[ht!]
  \centering
  \caption{Condition and disposition accuracy by type of query: comparison of a single evaluation run between a general reasoning model and our in-domain approach, with $k=5$ retrieved documents.}\label{tbl:error_analysis}
  \begin{tabular}{@{}lcrr@{}}
    \toprule
    \textbf{Type of Query} & \textbf{Model} & \textbf{Condition} & \textbf{Disposition} 
    \\
    \midrule
    Basic & t0-1.1-k5-32B  & 0.52 & \textbf{0.64} 
    \\
                           & s1.1-32B  & 0.49 & 0.52 
    \\[1ex]
    Hypochondriac &  t0-1.1-k5-32B  & \textbf{0.62} & 0.44 
    \\
                           & s1.1-32B  & 0.52 & 0.37 
    \\[1ex]
    Downplay &  t0-1.1-k5-32B  & 0.53 & 0.47 
    \\
                           & s1.1-32B  & 0.47 & 0.43 
    \\
    \bottomrule
  \end{tabular}
\end{table}

Overall, having a lean reasoning model trained on reasoning traces that show how to tackle decisions relevant to the domain under study, brings a clear advantage compared to a more general purpose reasoner. While the number of traces needed to train such in-domain model is not excessive (the \texttt{s1} study adopted one $1\,000$ traces, whereas we have used $2\,000$), obtaining them remains a challenge for applications focusing on private or sensitive collection of data. To address this problem, several strategies could be considered, such as:
\begin{itemize}
    \item manually creating reasoning traces based on a set of given queries and retrieved documents;
    \item hosting and querying a general-purpose reasoning model on local or private infrastructure, with manual correction of its responses to generate the needed traces;
    \item taking a general-purpose reasoner and further fine-tuning it on local or private infrastructure (which may require even fewer traces).
\end{itemize}

While these options are more complex to implement compared to relying on a frontier reasoning model, they do not constitute a fundamental barrier to adopting lean retrieval augmented reasoning models in secure environments or on sensitive data.

\subsection{Distillation to further reduce model size}

In our work, we have followed the strategy recently presented by the \texttt{s1} team \cite{muennighoff2025s1}, to distil reasoning capabilities from a frontier model such as \texttt{DeepSeek-R1} to a relatively small model such as our 32B parameters solution. While our model outperforms all other lean approaches for the task, as discussed in Tables \ref{tbl:llm_accuracy} and \ref{tbl:error_analysis}, it still requires around 64GBs of GPU memory to load the model in 16-bit precision. As such request is not always readily available for teams which are employing secure environments with resource constraints, in this section we explore whether we can further reduce the model size without significantly impacting the overall performance. The possibility of enhancing reasoning abilities in small language models via distillation is one of the main contributions of \cite{guo2025deepseek}. However as shown in Figure \ref{fig:model-performance}, depending on the task, the model will start outperforming its non-reasoning baseline only from a certain size. In a similar fashion, we evaluate the performance of distilled models ranging from 1.5B to 32B parameters, with the smallest models requiring only 3 to 6~GB of GPU memory, making them suitable for execution on most modern laptops\footnote{All our distilled models are available at \url{https://huggingface.co/alan-turing-institute}.}.

\begin{table}[ht!]
  \centering
  \caption{Condition and disposition accuracy with different model sizes, with $5$ documents retrieved. We also report as reference the performance of the 32B non reasoning baseline (\texttt{Qwen2.5-32B-Instruct}) and of the frontier model from which reasoning was distilled from (\texttt{DeepSeek-R1}), with the same number of documents retrieved.}\label{tbl:model_sizes}
  \begin{tabular}{@{}lrrr@{}}
        \textbf{Model Size} & \textbf{Memory (GB)} & \textbf{Condition} & \textbf{Disposition} 
    \\
    \midrule
    1.5B & 3 & 0.53 & 0.47 
    \\
    3B & 6 & \textbf{0.56} & 0.48 
    \\
    7B & 14 & 0.54 & 0.48 
    \\
    14B & 28 & \textbf{0.56} & 0.48 
    \\
    32B & 64 & \textbf{0.56} & \textbf{0.51} 
    \\[1ex]
    32B (Qwen baseline) & 64 & 0.54 & 0.50\\ 
    671B (DeepSeek-R1) & $1\,342$ & \textbf{0.56} & \textbf{0.51} 
    \\
    \bottomrule
  \end{tabular}
\end{table}

\begin{figure}[htbp]
  \centering
  \includegraphics[width=\linewidth]{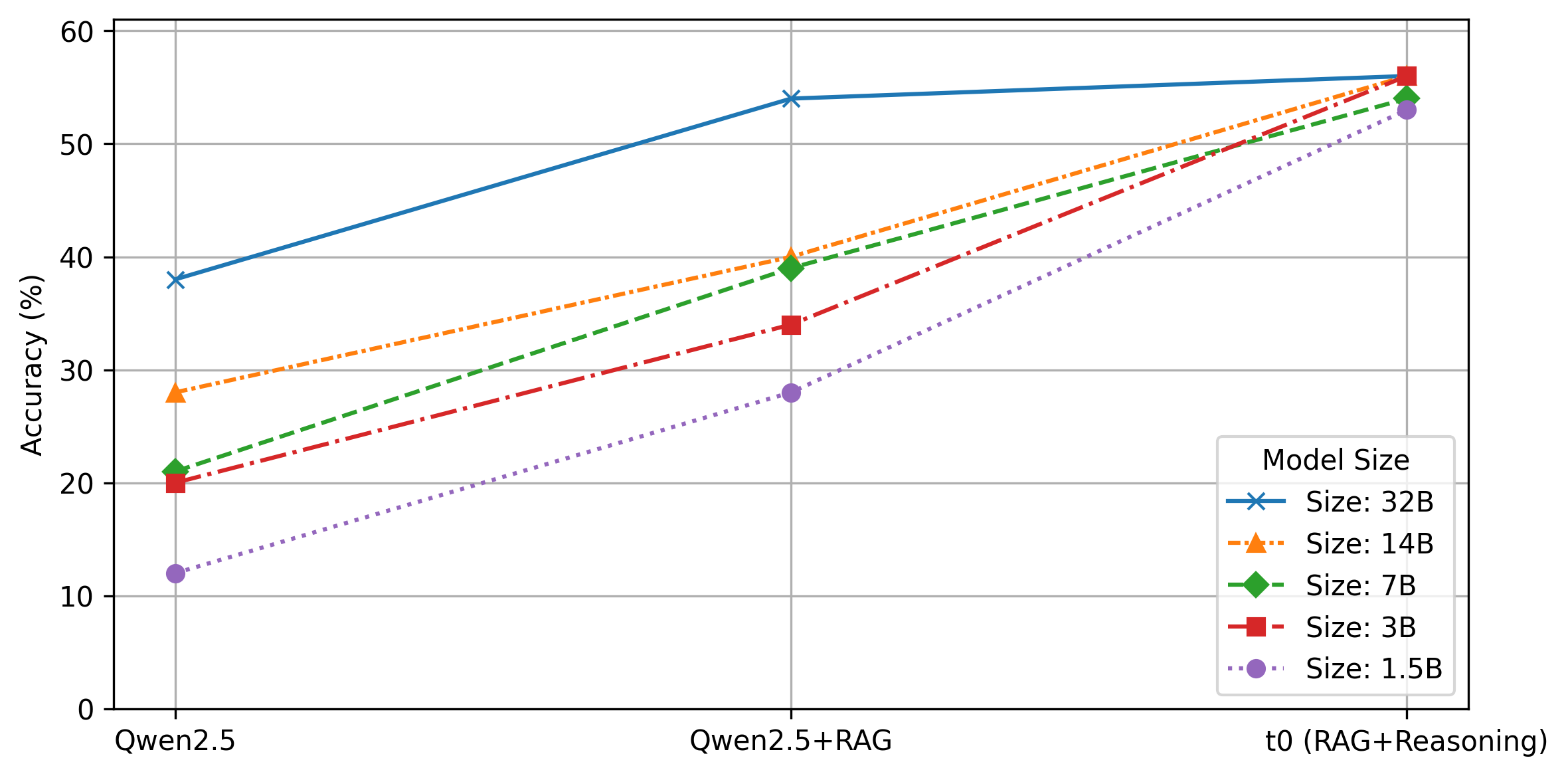}
  \caption{Performance on condition prediction of \texttt{Qwen2.5-Instruct} models alone, with the use of RAG and with their post-trained \texttt{t0} versions which combine RAG and reasoning.}
  \label{fig:lean-model-comparison}
\end{figure}

As shown in Table~\ref{tbl:model_sizes},\footnote{We compute a rough estimate of the memory required to store the model parameters using a 16-bit precision using the formula: $\text{GPU memory needed (bytes)} = \text{2} \times \text{Number of parameters}$, as $2$ bytes are needed per parameter.} distilling reasoning capabilities into smaller models is feasible: even a 1.5B parameter model maintains performance comparable to a 32B parameter non-reasoning model, especially in condition prediction, and learns effectively from its 671B parameter frontier ``teacher.'' To better understand the performance boost that the combination of RAG and reasoning offers at different model sizes, we have presented in Figure~\ref{fig:lean-model-comparison} a comparison for condition prediction between each initial \texttt{Qwen2.5-Instruct} model, the same model with RAG and finally the post-trained \texttt{t0} version, which combines RAG and reasoning. The figure highlights an important insight for many downstream applications: while for a 32B parameter model, the primary performance gain comes from the model's core ability to interpret the retrieved information, for leaner models (1.5B and 3B), a large performance boost is given by in-domain reasoning training which offers this interpretation ability that was otherwise lacking.\footnote{At size 1.5B for instance, \texttt{Qwen2.5+RAG} and \texttt{t0} are provided with the same documents retrieved for each query and the only difference is that the reasoning model \texttt{t0} is now provided with the ability to process the content in more details and therefore predicting the condition much better.} Such small-scale retrieval augmented reasoning models can in-fact deliver performance comparable to models over ten times larger, which significantly broadens the range of deployment scenarios. Such models are lightweight enough to run on many consumer-grade laptops, making them viable for widespread use in research and government applications. 

As extensively discussed in our report, if other teams want to pursue this, it is essential to set a clear benchmark for the desired task, to understand (i)~whether a combination of retrieval and reasoning could improve performance and reduce hallucinations and other types of errors, compared to a base model, and (ii)~what is the best trade-off between model size and performance, as depending on the application, a leaner solution could still deliver reliable performance while very significantly reducing the computational requirements and costs.

\subsection{The frontend interface} \label{sec:interfact}

\begin{figure}[t]
  \centering
  \includegraphics[width=\linewidth]{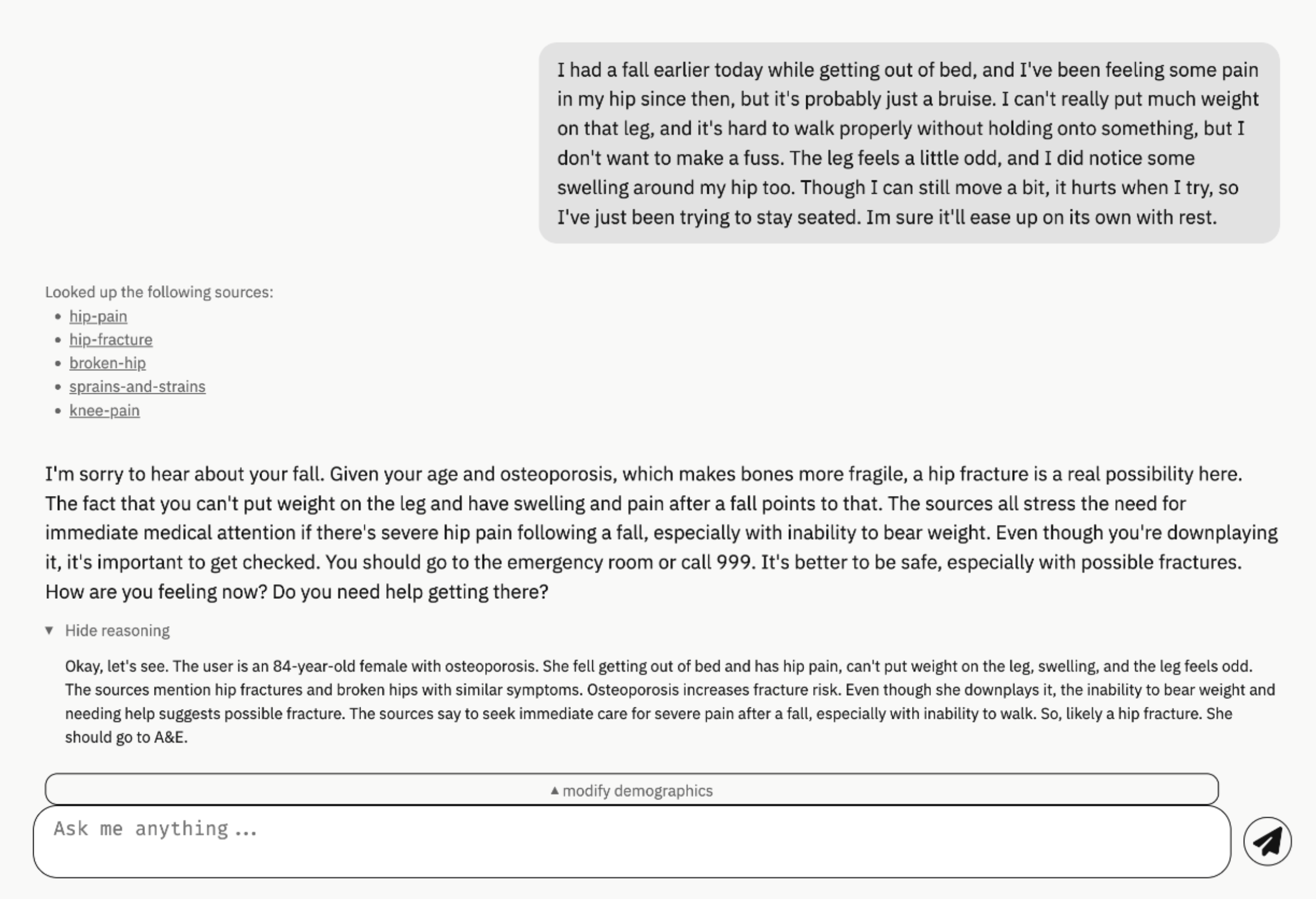} 
  \caption{Snapshot of the chat interface.}
  \label{fig:interface}
\end{figure}

In the GitHub repository, we provide a simple frontend interface
to see how our \texttt{t0-1.1-k5-32B} model could be used in practice, as
part of a larger system which handles model orchestration and
multi-turn chat interactions. In the snapshot of the interface
presented in Figure \ref{fig:interface}, the main components are
presented: given a user query (one of our synthetically generated
examples), \texttt{Qwen2.5-Instruct-32B} decides to invoke the
retriever, which collects $5$ relevant conditions (hip-pain,
hip-fracture, etc.). Then, \texttt{t0-1.1-k5-32B} generates a
reasoning process based on the query and the provided documents (shown in the snapshot using the drop-down), before answering to
the user suggesting one of the three possible dispositions (self-care, urgent primary care or as in this case A\&E)\footnote{During experimentations, we have also considered the fourth option ``request ambulance,'' which would have been predicted in this case to avoid risking a further fall by mobilising.}. The web frontend also includes a form that is used for
entering demographic information associated with the query.

The provided frontend can be seamlessly adapted for many other applications that would rely on combining model orchestration, reasoning capabilities and retrieval (with the possibility of adding additional metadata) on a collection of documents.


\section{Conclusion}

In this technical report, we have described how we have effectively combined reasoning and retrieval augmented generation together in a single lean model. To show its usefulness for decision making on domain-specific collections, we have presented a case study about determining conditions and disposition of a series of synthetic requests using the NHS A-to-Z collection as body of knowledge. Our model performed at comparable level with frontier reasoning model and especially outperformed other small-scale reasoning models that have been trained on mathematical reasoning and not fine-tuned for domain-specific applications. Finally, we highlight that it is possible to further reduce model size via distillation of reasoning capabilities in very small models, while keeping strong performance. We hope our overview and the paired GitHub codebase will be useful to others interested in combining reasoning and retrieval capabilities in domain-specific settings.

\section*{Acknowledgments}

RC, FN, and TL contributed equally to this work, leading respectively the implementation (RC), the overall project (FN) and the computational work (TL). Based on the CRediT taxonomy, these are the contributions of all authors: Conceptualisation (AD, JG, FN), Implementation (RC, TL, FN, RW, PY), Computational infrastructure (RC, TL, RW), Data curation (RC, JG, FN, RW), frontend interface (PY), Original draft (RC, FN, TL), Reviewing \& Editing (all), Advising (AD, JG, MG, LT), Project management (AD, JG, FN).

This work was funded by The Alan Turing Institute. We would like to thank Christopher Banerji, Maya Bronfeld, Jonathan Carter, Tom Jeffery and Giles Lawrence for their invaluable support and constructive feedback throughout the course of the project.

The computations described in the report were in part performed using the Baskerville\footnote{\url{https://www.baskerville.ac.uk/}} Tier 2 HPC service. Baskerville was funded by the EPSRC and UKRI through the World Class Labs scheme (EP/T022221/1) and the Digital Research Infrastructure programme (EP/W032244/1) and is operated by Advanced Research Computing at the University of Birmingham.

The authors also acknowledge the use of resources provided by the Isambard-AI National AI Research Resource (AIRR). Isambard-AI is operated by the University of Bristol and is funded by the UK Government's Department for Science, Innovation and Technology (DSIT) via UK Research and Innovation; and the Science and Technology Facilities Council [ST/AIRR/I-A-I/1023].

\bibliographystyle{unsrt}
\bibliography{ref}

\appendix

\section{Appendix}

\subsection{Conversational RAG system prompt template} \label{app:rag_conv_template}

For our conversational RAG pipeline described in Section \ref{sec:rag_conv}, below is the system prompt template we used for the language model to generate a response when retrieval is used.

\begin{lstlisting}
You are a helpful clinical AI assistant deployed in the United Kingdom

You will be given a description of some of the users symptoms and some retieved context from NHS condition web pages which provide information about various medical conditions that could be relevant to those symptoms.

Use the description of the users symptoms, the following retrieved context and similarity scores for each piece of context (a lower similarity score means the higher similarity to the patient's query) to work out what condition(s) the user is suffering from and provide a recommendation of what they should do next.
Never state or refer to the similarity scores to the user.

Ask follow up questions to the user to gather more information or for further details about their symptoms to narrow down the potential conditions.
Focus on the most serious conditions first.

In your response, reply in English and always refer to the user in the second person.

If you don't know the answer to a question, just say that you don't know.
If the retrieved context is not relevant to the patient's query, you should also say that you don't know.

Retrieved context:
{context}

This is a summary of their demographics:
{demographics}
\end{lstlisting}

\subsection{Conversational agent system prompt} \label{app:conv_template}

For our conversational RAG pipeline described in Section \ref{sec:rag_conv}, below is the system prompt to the conversational agent language model which decides whether to use the retrieval tool or to reply directly. 

\begin{lstlisting}
You are a helpful clinical AI assistant deployed in the United Kingdom

You are provided a tool that can retrieve context from a knowledge base taken from NHS condition web pages which provide information about various medical conditions.
You should always use the tool to find relevant information to answer the patient's question rather than relying on your own knowledge.
If you are confused or unsure about the user's question, you should use the tool to find relevant information or ask the user for more information or ask further details about their symptoms.
For follow up questions from the user, you should always use the tool to find new relevant information to answer the user's question given the conversation history.
You should only not use the tool in very simple messages that do not require any context like "Hello" or "Thank you", or when the user is just writing something random.

You can also ask the user for more information or ask further details about their symptoms.
If you are going to reply to the user, always conclude with a question to keep the conversation going to help the user or ask for more details about their symptoms.
In your response, only reply in English and always refer to the user in the second person.

Decide to use the tool at the start. Do not use the tool after you have already started your response.
\end{lstlisting}

\subsection{Prompt template for summarisation} \label{app:summarisation_prompt}

For obtaining summarised versions of our documents as described in Section \ref{sec:nhs_dataset}, we used the following user prompt:

\begin{lstlisting}
Summarise the document below, focusing only on symptoms and how to decide the next course of action. Be concise - aim for a summary of 3-4 sentences or fewer, keeping only essential information.

Document:
{document}
\end{lstlisting}

\subsection{Prompt template for synthetic user query generation}
\label{app:synthetic_data_prompt}

For generating synthetic user queries as discussed in Section \ref{sec:eval_synthetic}, the following user prompt was used:

\begin{lstlisting}
Generate a synthetic NHS 111 query based on the following details:

### Query Type:
* "basic": Based on a single condition page, the query mentions relevant symptoms
* "hypochondriac": Based on a single condition page, the query mentions relevant symptoms plus other unrelated complaints and expressions of excessive anxiety
* "downplay": Based on a single condition page, the query downplays the severity of the symptoms

### Condition Content Source:
* The primary textual content extracted from the relevant NHS condition web pages

### Severity Level:
* A&E: Emergency hospital treatment required
* Urgent Primary Care: patient should be seen as soon as possible, by a GP, urgent care centre, or similar
* Self-care: Issue can be handled at home and/or with over-the-counter medication

### Required JSON Output:
Return the query in the following structured JSON format:

```json
{
  "general_demographics": {
    "age": "[Realistic adult age given symptoms and severity, e.g., 20-80, for anyone above 80 use 'above 80']",
    "sex": "{sex}",
    "occupation": "[A common occupation]",
    "social_support": "[Specify if the patient has a social support network, such as a partner, family member, or living carer. If applicable, include details like the carer's role (e.g., 'My partner is here to help me' or 'I live with my daughter who is my carer'). If no support network is present, state 'No support network.']",
    "medical_history": "[Include any relevant comorbidities, such as diabetes, asthma, neurodegenerative conditions (e.g., Alzheimer's, Parkinson's), allergies (e.g., to medications, food, or environmental triggers), or other significant pre-existing health conditions. If the person is on regular medications (e.g., insulin for diabetes, inhalers for asthma, antihistamines for allergies, etc.), list them as well. If there are no significant conditions, medications, or allergies, keep it simple (e.g., 'No known issues' or 'None relevant'). Only include specific conditions, medications, or allergies if they are highly relevant to the current case or commonly co-occur with the condition in question.]"
  },
  "symptoms_description": "[Generate a natural-sounding, first-person query (using 'I', 'my') as if a patient is describing their symptoms to NHS 111. Ensure the described symptoms are primarily drawn from or plausibly related to the condition content AND strongly align with the specified severity_level. Select/adapt details from condition content justifying the target severity (e.g., 'red flag' symptoms for Urgent Primary Care; milder symptoms for Self-care). Ensure consistency with the query_type. Vary tone (e.g., anxious, calm) and sentence structure for realism. Occasionally include precise details, such as temperature readings or numbers from previous exams (e.g., 'My temperature is 39C or 102F'). At other times, be vague when describing symptoms (e.g., 'I have a high temperature'). Numbers can be in either US or UK format, depending on the context.]"
}
```

Ensure the text is realistic, clear, and varied to simulate real-world queries.

If the provided conditions content lacks sufficient symptom information or actionable guidance to construct a plausible scenario for the requested severity_level, please return:
{"error": "Insufficient symptom information in provided content to match requested severity"}

Example Output (Urgent Primary Care, Basic Query for Headache):

```json
{
  "general_demographics": {
    "age": 35,
    "sex": "Female",
    "occupation": "Teacher",
    "social_support": "No support network",
    "medical_history": "No known chronic conditions"
  },
  "symptoms_description": "I've had a severe headache for the past three days that won't go away, even with painkillers. It feels like a tight band around my head, and I'm also feeling slightly nauseous. My vision is a bit blurry when I stand up too quickly. I don't normally get headaches this bad, and I'm starting to feel concerned."
}
```

Reply only with the JSON output, without any additional text or explanation

Generate a query based on the given details:
Query Type: {query_type}
Severity Level: {severity_level}
Sex: {sex}
Conditions web page content: {conditions_content}
\end{lstlisting}

\subsection{Prompt template for reasoning traces generation} \label{app:reasoning_traces_prompt}

To generate reasoning traces from \texttt{DeepSeek-R1} to predict the condition and severity of the query as described in Section \ref{sec:fine_tuning_process}, we used the following prompt template:

\begin{lstlisting}
Use the following pieces of retrieved context and similarity scores (lower scores means higher similarity to the patient's query):
{context}

A patient has given the following description of their symptoms:
"{question}"

This is a summary of their demographics:
{demographics}

Using the sources and context provided, submit the condition and the severity level in the format: "(condition, severity)". Do not provide any explanation to the output, only your final answer.

Remember that the condition must either be one of {sources} or "inconclusive" if you think that the condition is not listed.
Remember that the severity level must be one of ["Self-care", "Urgent Primary Care", "A&E"].
\end{lstlisting}

In this template, we present to the model the retrieved context from our retriever along with their similarity scores (L2-norm), the user query and their demographics (which are synthetically generated as per Section \ref{sec:eval_synthetic}) and finally the titles of the documents retrieved (i.e., the retrieved conditions).

\subsection{Prompt templates for condition and next action prediction} \label{app:cond_next_action_pred_prompt}

In this section, we share the prompts used to evaluate various models on the task of determining the condition of a synthetic patient and the next course of action as outlined in Section \ref{sec:cond_next_action_pred}. For several models considered (\texttt{Qwen2.5-32B-Instruct}, \texttt{GPT-4o}, \texttt{o3-mini}, \texttt{Qwen3-32B}), to obtain a condition and next action prediction, we presented the model with a tool to submit their answer which was simply a function with two arguments: condition and severity. This ensured simple and consistent parsing of the model predictions. The system and user prompts for these models are presented in Appendix \ref{app:cond_next_action_pred_prompt_tool_use}. For details on \emph{tool binding}, the process whereby models are made aware of available tools, see LangChain's documentation on tool calling at \url{https://python.langchain.com/docs/concepts/tool_calling/}.

However, for other models (\texttt{DeepSeek-R1}\footnote{At time of evaluation, tool-use and function-calling was not supported for \texttt{DeepSeek-R1} on Azure AI Foundry.}, \texttt{s1.1-32B}, \texttt{t0-1.1-k5-32B}), tool-use was not available. For these models, we used a different prompt to ask the model to submit the condition and the severity level prediction directly in the format: \texttt{"(condition, severity)"}. We wrote parsers to obtain the condition and severity levels. The system and user prompts for these models are presented in Appendix \ref{app:cond_next_action_pred_prompt_no_tool_use}.

Lastly, we also evaluate model performance without the retrieval system. In this case, we just present the model the list of $989$ conditions. The prompts used for models with and without tool-use options are presented in Appendix \ref{app:cond_next_action_pred_prompt_no_context_tool_use} and \ref{app:cond_next_action_pred_prompt_no_context_no_tool_use}, respectively. For brevity, we do not add the full list of $989$ conditions here.

\subsubsection{Prompt templates for models with tool-use options} \label{app:cond_next_action_pred_prompt_tool_use}

\textbf{System prompt}

\begin{lstlisting}
You are a clinical AI assistant.

You will be given a description of a patient's symptoms, some retrieved context that could be relevant to those symptoms and similarity scores for each piece of context (a lower similarity score means the higher similarity to the patient's query).

You need to suggest the most likely condition and the level of severity.

You should decide one of these options for severity:
* A&E: Emergency hospital treatment required
* Urgent Primary Care: patient should be seen as soon as possible, by a GP, urgent care centre, or similar
* Self-care: Issue can be handled at home and/or with over-the-counter medication.

You should use the provided tool to submit the condition and severity level.

Importantly, if you think that the condition is not listed, please use "inconclusive" for the condition.
\end{lstlisting}

\smallskip\noindent\textbf{User prompt template}

\begin{lstlisting}
Use the following pieces of retrieved context and similarity scores (lower scores means higher similarity to the patient's query):
{context}

A patient has given the following description of their symptoms:
"{question}"

This is a summary of their demographics:
{demographics}

Using the sources and context provided, use the "submit_condition_recommendation" tool to submit the condition and the severity level.

Remember that the condition must either be one of {sources} or "inconclusive" if you think that the condition is not listed.
Remember that the severity level must be one of ["Self-care", "Urgent Primary Care", "A&E"].
\end{lstlisting}

\subsubsection{Prompt templates for models without tool-use options} \label{app:cond_next_action_pred_prompt_no_tool_use}

\textbf{System prompt}

\begin{lstlisting}
You are a clinical AI assistant.

You will be given a description of a patient's symptoms, some retrieved context that could be relevant to those symptoms and similarity scores for each piece of context (a lower similarity score means the higher similarity to the patient's query).

You need to suggest the most likely condition and the level of severity.

You should decide one of these options for severity:
* A&E: Emergency hospital treatment required
* Urgent Primary Care: patient should be seen as soon as possible, by a GP, urgent care centre, or similar
* Self-care: Issue can be handled at home and/or with over-the-counter medication.

Importantly, if you think that the condition is not listed, please use "inconclusive" for the condition.
\end{lstlisting}

\smallskip\noindent\textbf{User prompt template}

\begin{lstlisting}
Use the following pieces of retrieved context and similarity scores (lower scores means higher similarity to the patient's query):
{context}

A patient has given the following description of their symptoms:
"{question}"

This is a summary of their demographics:
{demographics}

Using the sources and context provided, submit the condition and the severity level in the format: "(condition, severity)". Do not provide any explanation to the output, only your final answer.

Remember that the condition must either be one of {sources} or "inconclusive" if you think that the condition is not listed.
Remember that the severity level must be one of ["Self-care", "Urgent Primary Care", "A&E"].
\end{lstlisting}

\subsubsection{Prompt templates without retrieved context for models with tool-use options} \label{app:cond_next_action_pred_prompt_no_context_tool_use}

\textbf{System prompt}

\begin{lstlisting}
You are a clinical AI assistant.

You will be given a description of their symptoms.

You need to suggest the most likely condition and the level of severity.

You should decide one of these options for severity:
* A&E: Emergency hospital treatment required
* Urgent Primary Care: patient should be seen as soon as possible, by a GP, urgent care centre, or similar
* Self-care: Issue can be handled at home and/or with over-the-counter medication.

You should use the provided tool to submit the condition and severity level.

Importantly, if you think that the condition is not listed, please use "inconclusive" for the condition.
\end{lstlisting}

\smallskip\noindent\textbf{User prompt template}

\begin{lstlisting}
Use the following list of possible conditions:
[...]

A patient has given the following description of their symptoms:
"{question}"

This is a summary of their demographics:
{demographics}

Using the sources provided, use the "submit_condition_recommendation" tool to submit the condition and the severity level.

Remember that the condition must either be one of the conditions listed above or "inconclusive" if you think that the condition is not listed.
Remember that the severity level must be one of ["Self-care", "Urgent Primary Care", "A&E"].
\end{lstlisting}

\subsubsection{Prompt templates without retrieved context for models without tool-use options} \label{app:cond_next_action_pred_prompt_no_context_no_tool_use}

\textbf{System prompt}

\begin{lstlisting}
You are a clinical AI assistant.

You will be given a description of their symptoms.

You need to suggest the most likely condition and the level of severity.

You should decide one of these options for severity:
* A&E: Emergency hospital treatment required
* Urgent Primary Care: patient should be seen as soon as possible, by a GP, urgent care centre, or similar
* Self-care: Issue can be handled at home and/or with over-the-counter medication.

Importantly, if you think that the condition is not listed, please use "inconclusive" for the condition.
\end{lstlisting}

\smallskip\noindent\textbf{User prompt template}

\begin{lstlisting}
Use the following list of possible conditions:
[...]

A patient has given the following description of their symptoms:
"{question}"

This is a summary of their demographics:
{demographics}

Using the sources and context provided, submit the condition and the severity level in the format: "(condition, severity)". Do not provide any explanation to the output, only your final answer.

Remember that the condition must either be one of the conditions listed above or "inconclusive" if you think that the condition is not listed.
Remember that the severity level must be one of ["Self-care", "Urgent Primary Care", "A&E"].
\end{lstlisting}

\end{document}